\documentclass[review]{SCIS2021}
\usepackage{cite}

\usepackage{enumitem}
\newcommand{\cmark}{\ding{51}}%
\newcommand{\xmark}{\ding{55}}

\begin{document}

\ArticleType{ }
\Year{2020}
\Month{}
\Vol{}
\No{}
\DOI{}
\ArtNo{}
\ReceiveDate{}
\ReviseDate{}
\AcceptDate{}
\OnlineDate{}

\title{Densely Nested Top-Down Flows for Salient Object Detection}{Densely Nested Top-Down Flows}

\author[1$\dagger$]{Chaowei Fang}{}
\author[2$\dagger$]{Haibin Tian}{}
\author[2$\ast$]{Dingwen Zhang}{{zhangdingwen2006yyy@gmail.com}}
\author[2]{Qiang Zhang}{}
\author[3]{Jungong Han}{}
\author[4]{Junwei Han}{}
\AuthorMark{Chaowei Fang}


\contributions{Chaowei Fang and Haibin Tian have the same contribution to this work.}

\address[1]{School of Artificial Intelligence, Xidian University, Xi'an {\rm 710071}, China}
\address[2]{School of Mechano-Electronic Engineering, Xidian University, Xi'an {\rm 710071}, China}
\address[3]{Department of Computer Science, Aberystwyth University, Aberystwyth, Wales}
\address[4]{Brain and Artificial Intelligence Laboratory, School of Automation, Northwestern Polytechnical University, Xi'an {\rm 710072}, China}

\abstract{With the goal of identifying pixel-wise salient object regions from each input image, salient object detection (SOD) has been receiving great attention in recent years. One kind of mainstream SOD methods is formed by a bottom-up feature encoding procedure and a top-down information decoding procedure. While numerous approaches have explored the bottom-up feature extraction for this task, the design on top-down flows still remains under-studied. To this end, this paper revisits the role of top-down modeling in salient object detection and designs a novel densely nested top-down flows (DNTDF)-based framework. In every stage of DNTDF, features from higher levels are read in via the progressive compression shortcut paths (PCSP). The notable characteristics of our proposed method are as follows. 
1) The propagation of high-level features which usually have relatively strong semantic information is enhanced in the decoding procedure; 2) With the help of PCSP, the gradient vanishing issues caused by non-linear operations in top-down information flows can be alleviated.
3) Thanks to the full exploration of high-level features, the decoding process of our method is relatively memory efficient compared against those of existing methods.
Integrating DNTDF with EfficientNet, we construct a highly light-weighted SOD model, with very low computational complexity.
To demonstrate the effectiveness of the proposed model, comprehensive experiments are conducted on six widely-used benchmark datasets. The comparisons to the most state-of-the-art methods as well as the carefully-designed baseline models verify our insights on the top-down flow modeling for SOD. The code of this paper is available at https://github.com/new-stone-object/DNTD.}


\maketitle

\section{Introduction}

Salient object detection \cite{han2018advanced} aims at performing the pixel-level identification of the salient object region from an input image. Due to its wide-ranging applications in vision and multimedia community, such as object detection \cite{lin2017feature}, video object segmentation \cite{zhang2018spftn}, and weakly supervised object mining \cite{zhang2019leveraging}, numerous efforts have been made in recent years to develop effective and efficient deep salient object detection frameworks.

\begin{figure}[t]
\centering
\includegraphics[width=0.9\linewidth]{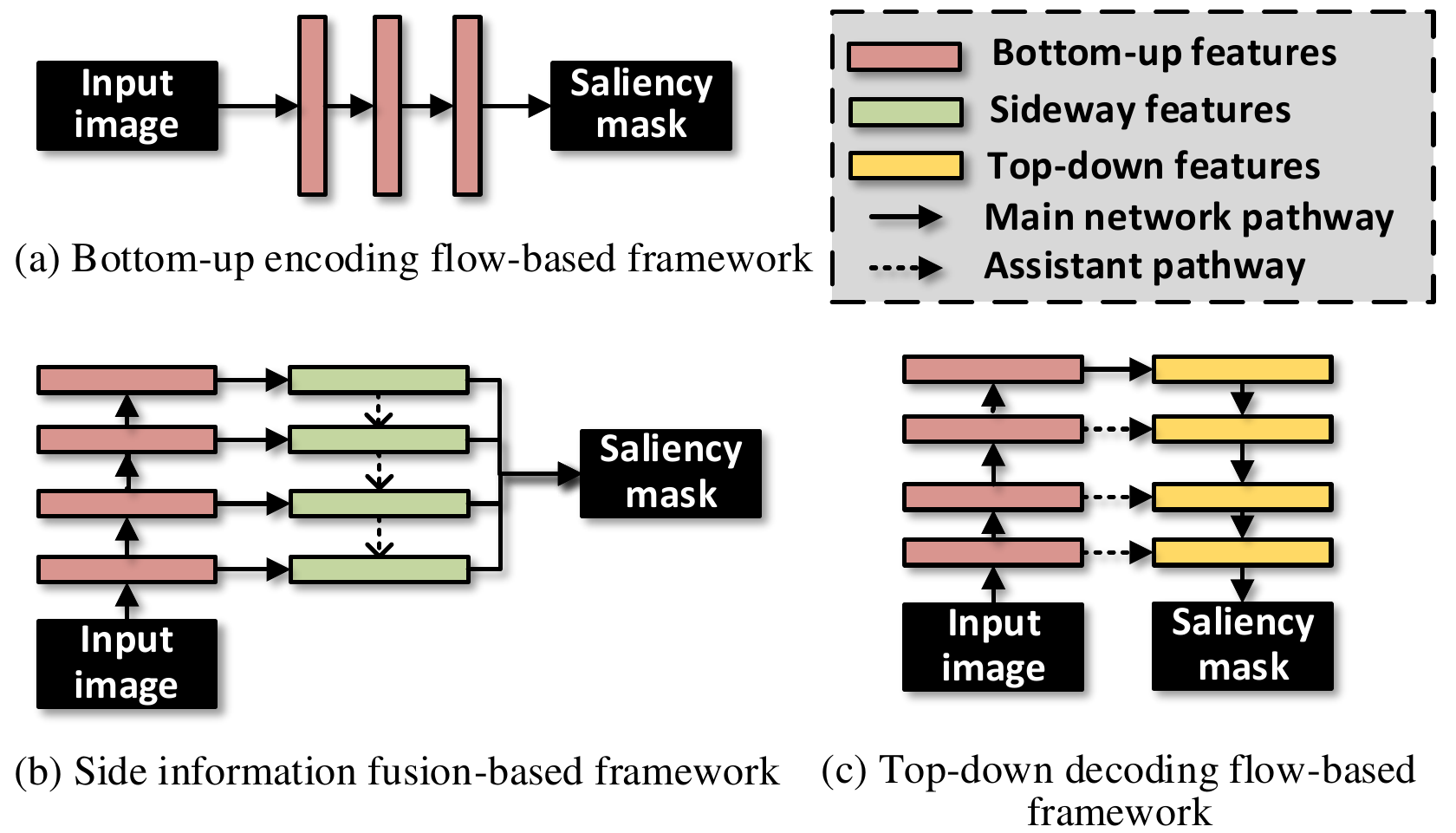}
   \caption{A brief illustration of the three mainstream designs of the deep salient object detection frameworks. In (a), the saliency map is derived from the topmost feature map of the backbone. 
   The kind of SOD methods in (b) attempt to explore multi-scale side information. (c) is the most popular decoding framework in SOD, which mines multi-scale features stage by stage.
     }
\label{fig:3designs}
\end{figure}

As shown in Figure \ref{fig:3designs}, the existing deep salient object detection models can be divided into three typical frameworks. The first one is the bottom-up encoding flow-based salient object detection framework  (see Figure \ref{fig:3designs} (a)). A bottom-up encoder is used for feature extraction, and then a simple classification head is attached on the top of the encoder for predicting the pixel-wise saliency map. Such methods \cite{liu2015predicting,li2015visual,zhao2015saliency,li2016deepsaliency,wang2016saliency} occur in relatively early ages in this research field by designing one or multiple forward network paths to predict the saliency maps. To take advantage of multi-stage feature representations, some recent works \cite{hou2017deeply,zhao2019pyramid,wu2019stacked,su2019selectivity,gao2020highly} start to incorporate additional network blocks to further explore the side information residing in the features extracted by multiple stages of the forward pathway. The involvement of the learned side information plays a key role in predicting the desired salient object regions. These works form the second type of learning framework, i.e., the side information fusion-based salient object detection framework (see Figure \ref{fig:3designs} (b)). Although the side information fusion-based frameworks have achieved great performance gains when compared to the bottom-up encoding flow-based frameworks, one important cue for saliency detection, i.e., the top-down information, has not been adequately explored. To this end, the third type of salient object detection framework occurred, which is named as the top-down decoding flow-based appeared (see Figure \ref{fig:3designs} (c)). In this framework, the main network pathway is formed by an encoder-decoder architecture, where the decoder explores saliency patterns from the multi-scale semantic embeddings stage by stage and gradually enlarges the resolution of the coarse high-level feature map  \cite{GateNet,MINet-CVPR2020,liu2018picanet,Feng_2019_CVPR,Liu2019PoolSal,zhang2018bi}. Notice that this framework may also use the side information to assist the decoding process, but the final saliency masks are obtained from the last decoder stage rather than the fusion stage of the side features.

\begin{figure}[t]
\begin{center}
\includegraphics[width=0.93\columnwidth]{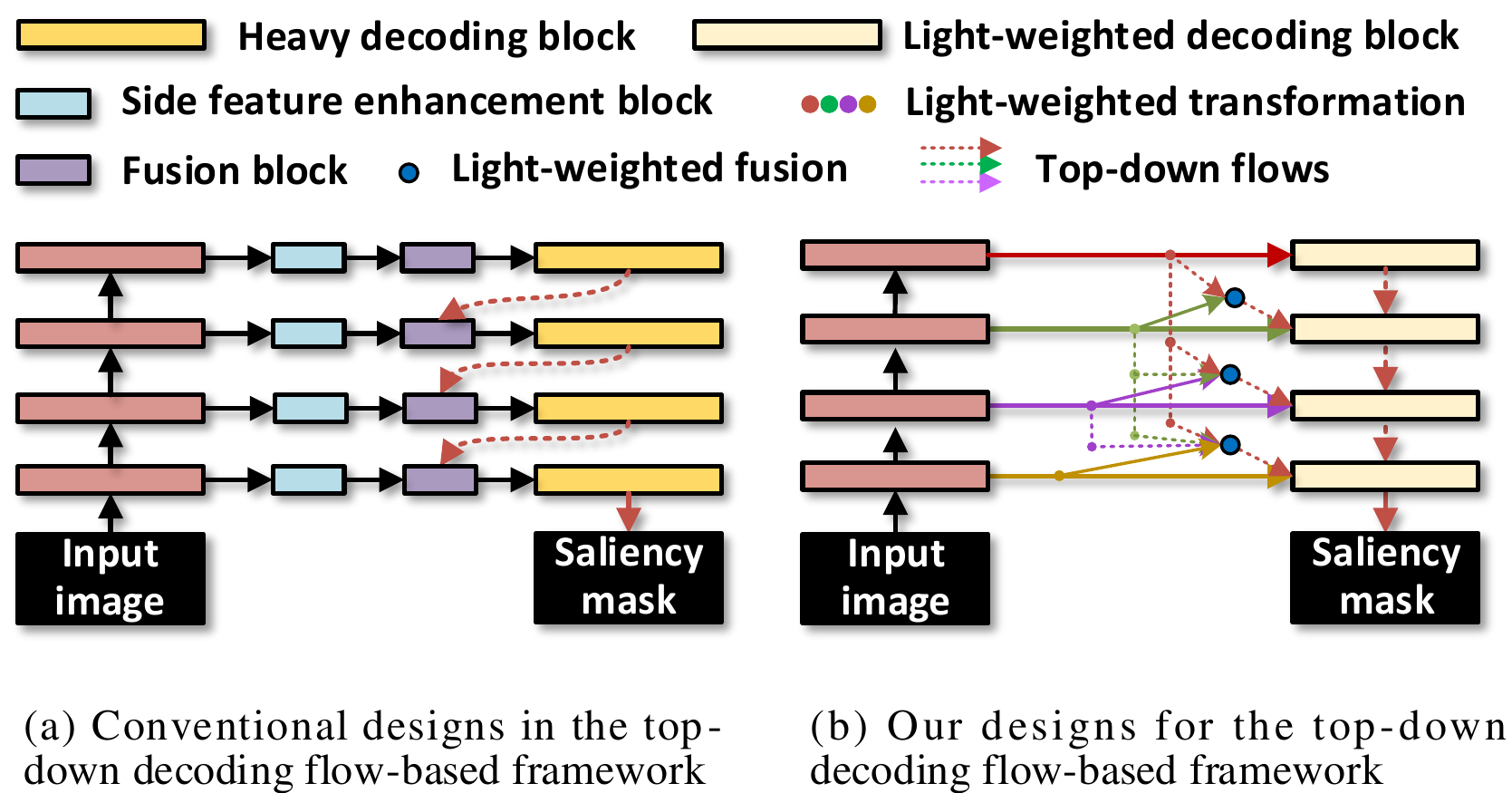}
\end{center}
   \vspace{11pt}
   \caption{Comparison between the conventional design and our design for the top-down decoding flow-based salient object detection framework.
   Our design explores richer semantic information from relatively top stages of the backbone during every decoding stage and complies with the light-weighted principle. 
     }
\label{fig:flow}
\end{figure}

From the aforementioned top-down decoding flow-based approaches, we observe that their core modeling components still focus on enhancing the side features and merging them into the decoding flow, whereas the top-down information flow remains primitive---propagating from the former decoding stage to the later one as is in the basic encoder-decoder architecture (see Figure \ref{fig:flow} (a)).
Considering that high-level features possess a great wealth of semantic information, we propose a novel decoding framework, named densely nested top-down flows (see Figure \ref{fig:flow} (b)), to enhance the exploration of features extracted from relatively higher levels.
In our method, feature maps obtained by each encoding stage are progressively compressed via shortcut paths and propagated to all subsequent decoding stages.
The strengthens of our method include the other two strong points. 1) The non-linear operations in the decoding stage are disadvantageous to the gradient back-propagation flow. Hence, the supervision signal propagated from the final prediction to the feature maps of top encoding levels might vanish. For example, if a neuron is not activated by the ReLU function, the gradient flow will be cut off, which means the supervision signal will not be propagated backward.
The progressive compression shortcut paths have no non-linear operations, hence they can relieve the gradient vanishing problem.
2) The reuse of high-level features allows a light-weighted decoding network design while achieving high salient object detection performance. Features produced by top layers of the encoder contain relatively strong semantic information which is beneficial to discriminate regions of salient objects from the background. Our method enhances the propagation of these features, resulting to a memory efficient decoding framework.





\begin{figure*}[t]
\begin{center}
\includegraphics[width=0.96\linewidth]{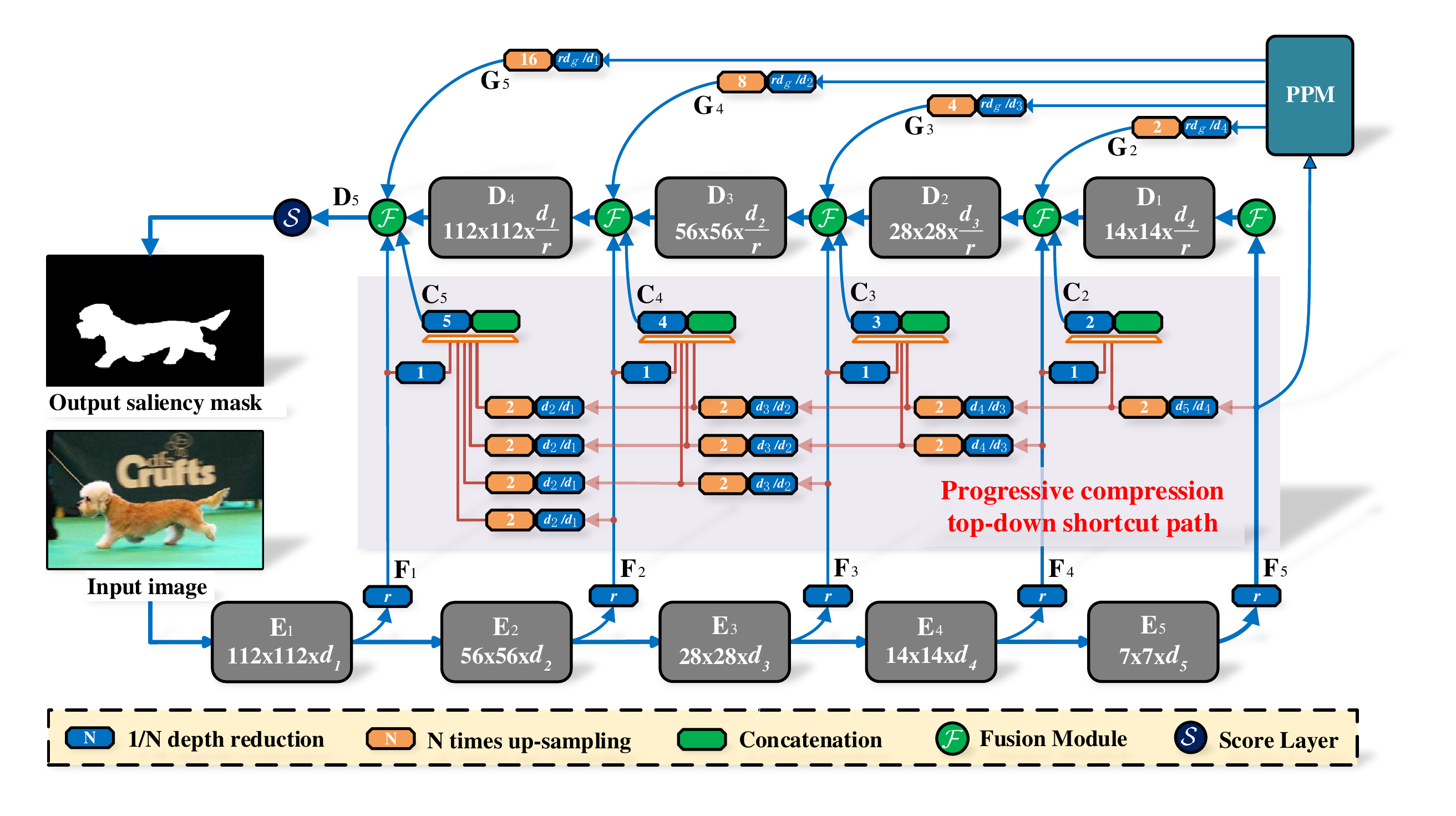}
\end{center}
   \caption{A brief illustration of the proposed salient object detection framework, which is built on a basic U-net-like architecture with the proposed DNTD to complement the rich top-down information into the decoding pathway. Inspired by \cite{Liu2019PoolSal}, the PPM is used to involve the global context features. The whole network is built by light-weighted network designs. More details of the network architecture can be referred to in Sec. 3.
     }
\label{fig:model_arch}
\end{figure*}

The overall framework is shown in Figure \ref{fig:model_arch}. As can be seen, we use the U-net-like architecture as the main network stream, upon which we further design a novel densely nested top-down flow path to introduce the rich top-down information to the decoding stages. To reduce the computational complexity of the decoding stages, we add a $1 \times 1$ channel compression layer to each side information pathway. In fact, all the feature pathways in the proposed decoding framework only need to pass through a number of $1 \times 1$ convolutional layers together with very small amounts of $3 \times 3$ convolutional layers, making the entire network highly lightweight. Meanwhile, the full exploration of top-down information in the proposed DNTD contributes to even better performance than the state-of-the-art SOD methods.

In summary, this work has the following three-fold main contributions: 1) We revisit an important yet under-studied issue, i.e., the top-down flow modeling, of the recent SOD frameworks; 2) We design a highly light-weighted learning framework, via introducing a novel densely nested top-down flow architecture; 3) Comprehensive experiments are conducted, demonstrating the promising detection capacity and the lower computational complexity of the proposed framework. Meanwhile, the insights on top-down modeling are well verified.

\section{Related Work}

Traditional salient object detection methods are designed based on the hand-crafted features~\cite{yang2013saliency,zhang2015minimum,cheng2014global,zhu2014saliency,jiang2013salient,klein2011center}. Recently, convolutional neural networks (CNN) have been extensively applied in salient object detection. Thanks to the powerful feature extraction capability of CNN, a great breakthrough has been made in devising effective SOD algorithms.
CNN-based SOD frameworks can be categorized into three kinds, including bottom-up encoding flow-based, side information fusion-based, and top-down decoding flow-based framework as shown in Figure \ref{fig:flow}.

In \textbf{bottom-up encoding flow-based framework}, one or multiple forward network pathes are designed to predict the saliency maps. For example, Liu \emph{et al}. \cite{liu2015predicting} propose a multi-resolution convolutional neural network, which has three bottom-up encoding pathways to deal with patches captured from three different resolutions of the input image. A similar idea is also proposed by Li and Yu \cite{li2015visual}, where a three-pathway framework is devised to extract multi-scale features for every super-pixel and two fully connected layers are adopted to fuse the features and predict the saliency value. In \cite{zhao2015saliency}, Zhao \emph{et al}. propose a multi-context deep learning framework, which fuses a global-context forward pathway and a local-context forward pathway to obtain the final prediction. In \cite{li2016deepsaliency}, Li \emph{et al}. build a multi-task deep learning model for salient object detection, where a shared bottom-up encoding pathway is used to extract useful deep features and two parallel prediction heads are followed to accomplish the semantic segmentation and saliency estimation, respectively. To learn to refine the saliency prediction results, Wang \emph{et al}. \cite{wang2016saliency} propose a recurrent fully convolutional network architecture. Specifically, they combine multiple encoding pathways, where saliency prediction results from the former encoding pathway are used to form the input of the latter one.

The \textbf{side information fusion-based salient object detection framework} aims to further explore the side information from the features extracted in each stage of the forward pathway.
Specifically, based on the network architecture of the holisitcally-nested edge detector \cite{xie2015holistically}, Hou \emph{et al}. \cite{hou2017deeply} introduce the skip-layer structures to provide rich multi-scale feature enhancement for exploring the side information. Zhao and Wu \cite{zhao2019pyramid} propose a simple but effective network architecture. They enhance the low-level and high-level side features in two sperate network streams, where the former is passed through a spatial attention-based stream while the later is passed through a channel attention-based stream. In \cite{wu2019stacked}, Wu \emph{et al}. also build a two-stream side information flow. However, different from \cite{zhao2019pyramid}, the two-stream side information flow is designed to fuse the multi-stage features for salient region identification and salient edge detection, respectively. In \cite{su2019selectivity}, Su \emph{et al}. use a boundary localization stream and a interior perception stream to explore different side features for obtaining the high-selectivity features and high-invariance features, respectively. Recently, Gao \emph{et al}. \cite{gao2020highly} propose gOctConv, a flexible convolutional module to efficiently transform and fuse both the intra-stage and cross-stage features for predicting the saliency maps.

To take advantage of top-down information, the third type of salient object detection framework emerges, i.e., the \textbf{top-down decoding flow-based framework}. In this framework, the main network pathway is formed by an encoder-decoder architecture, where the decoder recognizes out saliency patterns after fusing multi-scale features progressively. Notice that this framework may also uses the side information to assist the decoding process, but the final saliency masks are obtained from the last decoder stage instead of the fusion stage of the side features. One representative work is proposed by Zhang \emph{et al}. \cite{zhang2018bi}, where a U-net \cite{ronneberger2015u}-like architecture is used as the basic network and a bi-directional message passing model is introduced into the network to extract rich side features to help each decoding stage. Following this work, Liu \emph{et al}. \cite{Liu2019PoolSal} design a pooling-based U-shape architecture, where they introduce a global guidance module and a feature aggregation module to guide the top-down pathway. In \cite{Feng_2019_CVPR}, Feng \emph{et al}. propose an Attentive Feedback Module (AFM) and use it to better explore the structure of objects in the side information pathway. Liu \emph{et al}. \cite{liu2018picanet} propose the local attended decoding and global attended decoding schemes for exploring the pixel-wise context-aware attention for each decoding stage. More recently, in order to better explore the multi-level and multi-scale features, Pang \emph{et al}. \cite{MINet-CVPR2020} design aggregation interaction modules and self-interaction modules and insert them into the side pathway flow and decoding flow, respectively. In \cite{GateNet}, Zhao \emph{et al}. propose a gated decoding flow, where multi-level gate units are introduced in the side pathway to transmit informative context patterns to each decoding stage.
In this paper, we concentrate on further enhancing the usage of high-level features extracted by the encoder in the decoding flow.
A densely nested top-down flows-based decoding framework is proposed to encourage the reuse of high-level features in every stage of the decoding process. Compared to existing top-down decoding flow-based methods, the superiorities of our method are as follows. The gradient vanishing problem caused by the nonlinear operations in the decoding procedure can be mitigated, and a memory efficient decoding network is employed to facilitate the fusion of multi-stage features while maintaining high detection performance.

\section{Proposed Method}

The purpose of this paper is to settle the saliency object detection problem. Given an RGB image $\mathbf X$ with the size of $h\times w$, we propose a novel efficient deep convolutional architecture to predict a saliency map $\mathbf P\in[0,1]^{h\times w}$. Every element in $\mathbf P$ indicates the saliency probability value of the corresponding pixel.
A novel densely nested top-down flow architecture is built up to make full use of high-level feature maps.
The semantic information of top layers is propagated to bottom layers through progressive compression shortcut paths.
Furthermore, interesting insights are provided to design light-weighted deep convolutional neural networks for salient object detection. Technical details are illustrated in subsequent sections.

\subsection{Overview of Network Architecture}\label{sec:overview}

The overall network is built upon an encoder-decoder architecture, as shown in Figure \ref{fig:model_arch}.
After removing the fully connected layers, the backbone of an existing classification model, such as ResNet~\cite{he2016deep} and EfficientNet~\cite{tan2019efficientnet}, is regarded as the encoder.
Given an input image $\mathbf X$, the encoder is composed of five blocks of convolution layers, which yield 5 feature maps, $\{\mathbf E_i\}_{i=1}^5$.
Every block reduces the horizontal and vertical resolutions into half. Denote the height, width and depth of $\mathbf E_i$ be $w_i$, $h_i$ and $d_i$, respectively. We have, $h_{i+1}=h_i/2$ and $w_{i+1}=w_i/2$.

The target of the decoder is to infer the pixel-wise saliency
map from these feature maps. First of all, a compression unit is employed to reduce the depth of each scale of feature map,
\begin{equation} \label{eq:compress}
\mathbf F_i=\mathcal{C}_r(\mathbf E_i, \mathbf W_i^c),
\end{equation}
where $\mathcal{C}_r(\cdot,\cdot)$ indicates the calculation procedure of the depth compression unit, consisting of a ReLU layer~\cite{glorot2011deep} followed by a $1\times1$ convolution layer with the kernel of $\mathbf W_i^c$. $r$ represents the compression ratio, which means the depth of $\mathbf F_i$ is $d_i/r$.
Inspired from \cite{Liu2019PoolSal}, the pyramid pooling module (PPM)~\cite{he2015spatial,zhao2017pyramid} is used to extract a global context feature map $\mathbf G$ (with size of $h_5\times w_5\times d_g$) from the last scale of feature map $\mathbf F_5$ produced by the encoder.
Afterwards, a number of convolution layers are set up to fuse these compressed feature maps $\{\mathbf F_i\}_{i=1}^5$ and the global feature map $\mathbf G$, and output a soft saliency map, based on the U-shape architecture.
The distinguishing characteristics of our encoder are reflected in the following aspects:
1) In every stage of the decoder, the features of the top stages of the encoder are accumulated through progressive compression shortcut paths, forming into the feature representation for SOD together with the additional information learned in the current stage.
2) Our decoder is comprised of 1$\times$1 convolutions and a few 3$\times$3 convolutions, which only take up a small number of parameters and consume a small amount of computational cost.
The above decoder designs constitute our so-called densely-nested top-down flows.

\subsection{Densely Nested Top-Down Flow}\label{sec:denseflow}

In deep convolution neural networks, features extracted by top layers have strong high-level semantic information.
These features are advantageous at capturing the discriminative regions of salient objects.
Especially, when the network is pretrained on large-scale image recognition datasets, such as Imagenet~\cite{russakovsky2015imagenet}, the top feature maps are intrinsically capable of identifying out salient foreground objects according to~\cite{zhou2016learning}.
However, their spatial resolutions are usually very coarse which means it is difficult to locate fine object boundaries from them.
On the other hand, bottom layers produce responses to local textural patterns which is beneficial to locate the boundaries of the salient object.
Multi-resolution CNN models~\cite{li2015visual,liu2015predicting} uses multi-scale copies of the input image to explore both low-level and high-level context information. However, such kind of methods are usually cumbersome and cost heavy computation burden.
Inspired by the holistic-nested network architecture~\cite{xie2015holistically}, fusing multi-scale feature maps produced by different convolution blocks of the encoder is the other popular choice in SOD~\cite{LiYu16,hou2017deeply}. 
U-Net~\cite{ronneberger2015u}, which currently prevails in deep SOD methods~\cite{zhang2018bi,Liu2019PoolSal,MINet-CVPR2020,GateNet},  accumulates multi-scale feature maps in a more elegant manner. As shown in Figure \ref{fig:3designs}(c), the decoder usually shares the same number of stages with the encoder, and every stage in the decoder merges the feature map of the corresponding stage of the encoder forming a U-shape architecture.
However, in a standard U-Net, the features produced by the encoder are fused into the decoder via a simple linear layer. There exists room for improvement in more fully utilizing these features, especially these relatively high-level features.
The difficulty for propagating gradients back into the topper layers of the encoder increases as the gradient back propagation process needs to pass through more decoding stages.

For purpose of settling the above issues, we propose a novel top-down flow-based framework, named densely nested top-down flows. Shortcuts are incorporated to feed all feature maps of higher stages into every stage of the decoder.
The uniqueness of these shortcuts is that high-level feature maps are progressively compressed and enlarged stage by stage.
As shown in Figure \ref{fig:model_arch}, $\mathbf F_i$ is propagated to the bottom stages successively as follows,
\begin{eqnarray}
\nonumber \mathbf F_{i\rightarrow j} ={\mathcal Up}_{\times2}({\mathcal C}_{r_{j}^s}(\mathbf F_{i\rightarrow j-1}, \mathbf W_{i,j}^s)), \\
  \forall j, \; 5-i+2<j\leq5.
\end{eqnarray}
Here, $\mathbf F_{i\rightarrow j}$ indicates the semantic information propagated from $\mathbf F_i$ to $\mathbf F_j$. $\mathbf F_{i\rightarrow {5-i+2}}$($=\mathbf F_i$) is the initial input to the progressive compression shortcut path originated from $\mathbf F_i$. $\mathbf W_{i,j}^s$ represents weights of a 1$\times$1 convolution kernel, and the compression ratio $r_{j}^s$ is equal to $\frac{d_{5-j+1}}{d_{5-j+2}}$. ${\mathcal Up}_{\times2}(\cdot)$ upsamples the height and width of the input feature map into 2 times via the bilinear interpolation. These feature maps $\{\mathbf F_i^j\}_{i=5-j+2}^5$ generated by the progressive compression shortcuts are fed into the $j$-th ($j>1$) stage of the decoder.
No nonlinear function is used in the progressive compression shortcut path. Thus, the shortcut path can facilitate the gradient back-propagation, relieving the gradient vanishing issue caused by the multi-stage decoding process.
On the other hand, compared with reducing the depth into the target values at once, our progressive compression mechanism is more efficient, consuming less parameters.


The calculation process in the first stage of the decoder is a transition operation,
\begin{equation}
\mathbf D_1={\mathcal Up}_{\times2}(\mathcal{F}(\mathbf F_5, \mathbf W_{1}^d)),
\end{equation}
where $\mathcal{F}(\mathbf F_5, \mathbf W_{1}^d)$ consists of a ReLU layer and a $3\times3$ convolution layer with kernel of $\mathbf W$. It transmits $\mathbf F_5$ into a $h_4 \times w_4 \times d_4/r$ tensor defined as $\mathbf D_1$.

For subsequent stages in the decoder, the calculation process is composed of two fusion steps. First, for the $j$-th stage of the encoder,  we derive an additional feature map from the $(5-j+1)$-th stage of the encoder, $\hat{\mathbf F}_{5-j+1}=\mathcal C_1(\mathbf F_{5-j+1}, \mathbf W_{5-j+1}^a)$. The depths of the input and output feature maps are kept the same.
Together with $\hat{\mathbf F}_{5-j+1}$, the feature maps from higher-level stages are concatenated and compressed into a new context feature map,
\begin{equation}
\mathbf C_j=\mathcal{C}_j(\{\hat{\mathbf F}_{5-j+1},\mathbf F_{i\rightarrow j}|i=5-j+2,\cdots,5\}, \mathbf W_{j}^f\}).
\end{equation}
Note that the above fusion operation compresses the concatenated feature maps with the ratio of $j$, which indicates the depth of $\mathbf C_j$ is $d_{5-j+1}/r$.
The global feature $\mathbf G$ is complemented to the $j$-th stage of the decoder as well, $\mathbf G_j={\mathcal Up}_{\times2^{j-1}}(\mathcal C_{r^g_j}(\mathbf G, \mathbf W_j^g))$ where $r^g_j=\frac{d_g}{d_{5-j+1}}$.
Then, $\mathbf D_{j-1}$, $\mathbf F_{j}$, $\mathbf C_j$, and $\mathbf G_j$  are fused with a pre-placed ReLU and a $3\times3$ convolution layer, yielding the feature representation of the $j$-th stage of the decoder,
\begin{equation}
\mathbf D_j = {\mathcal Up}_{\times2}(\mathcal{F}(\{\mathbf D_{j-1}, \mathbf F_j, \mathbf C_j, \mathbf G_j\}, \mathbf W_{j}^d)).
\end{equation}
The depth of $\mathbf D_j$ is transformed into $d_{5-j}/r$. The final output is produced by a score prediction module consisting of a pre-placed ReLU layer, a $1\times1$ convolution layer and a Sigmoid function $\mathcal{S}(\cdot)$,
\begin{equation}
\mathbf P=\mathcal{S}({\mathcal Up}_{\times2}(\mathbf D_5, \mathbf W^o)),
\end{equation}
where $\mathbf W^o$ represents the kernel of the convolution layer, $\mathbf P$ ($h\times w\times 1$) is the final predicted saliency map.

The advantage of our densely nested decoder is that every stage is accessible to all higher-level feature maps of the encoder. This framework greatly improves the utilization of high-level features in the top-down information propagation flow. 

\subsection{Light-Weighted Network Design}\label{sec:light}

In this paper, we are not stacking piles of convolution layers to build a SOD network with high performance.
Our devised model has a light-weighted architecture while preserving high performance. 
	
Without backbones initialized with parameters pre-trained on Imagenet, it is difficult to achieve high performance via training a light-weighted backbone from scratch such as CSNet~\cite{gao2020highly}.
However, these initialized parameters are learned for solving the image recognition task. This means the features extracted by the pre-trained backbone are responsible for jointly locating the discriminative regions and predicting semantic categories.
In the SOD task, it is no longer necessary to recognize the category of the salient object. Considering the above point, we can assume that there exists a large amount of redundant information in the features extracted by the backbone. Hence, in our method, a large value is adopted for the compression ratio $r$ in (\ref{eq:compress}). We empirically find out that using $r\in\{2,4,8,16\}$ has little effects on the SOD performance in our method, as will be illustrated in the experimental section. With the help of a large compression ratio, the computation burden in the decoder can be greatly reduced.

Previous high-performance SOD models are usually equipped with decoders having a moderate amount of calculation complexity.
For example, \cite{he2015spatial} uses multiple $3\times3$ convolutions to construct a pyramid fusion module in every stage of the decoder. \cite{MINet-CVPR2020} adopts a number of $3\times3$ convolutions to aggregate inter-level and inter-layer feature maps in the decoder. Cascaded decoders are employed to implement top-down information in~\cite{F3Net,zhou2020interactive}, which lead to a decoding procedure with large computation burden.
In our proposed model, all convolutions adopted in the progressive compression shortcut paths have the kernel size of $1\times1$. This makes these complicated shortcuts only cost a few weights and computation resources in fact.
Furthermore, benefitted from rich top-down information, employing a single $3\times3$ convolution in every encoder stage is sufficient to construct a high-performance decoder.

The above network designs help us build up an effective and cost-efficient salient object detection model.

\begin{table}
\centering
\setlength{\tabcolsep}{0.5mm}{
\begin{tabular}{lcrr ccc ccc}
\toprule[1pt]
     &      &   &   & \multicolumn{3}{c}{DUTS-TE}  &     \multicolumn{3}{c}{HKU-IS}  \\
\cmidrule(l){5-7} \cmidrule(l){8-10}
\textbf{Model} & \textbf{Backbone} & \multicolumn{1}{c}{\textbf{Param}} & \multicolumn{1}{c}{\textbf{FLOPs}} & \textbf{$\text{F}_{\text{max}}$$\uparrow$} & \textbf{MAE$ \downarrow $} & \textbf{S$ \uparrow $} & \textbf{$\text{F}_{\text{max}}$$\uparrow$} & \textbf{MAE$ \downarrow $} & \textbf{S$ \uparrow $} \\
\midrule[1pt]

\multicolumn{10}{c}{\textbf{Resnet \& VGG}} \\
\midrule

\textbf{MLMSNet~\cite{wu2019mutual}} & VGG16    & 68.024M & 162.515G    & 0.854  & 0.048  & 0.861  & 0.922  & 0.039  & 0.907  \\
\textbf{BASNet~\cite{Qin_2019_CVPR}} & ResNet34 & 87.060M & 161.238G    & 0.860  & 0.047  & 0.866  & 0.929  & 0.032  & 0.909  \\
\textbf{CSF+Res2Net~\cite{hou2017deeply}} & Res2Net & 36.529M & 13.223G & 0.893  & 0.037  & 0.890  & 0.936  & 0.030  & 0.921    \\
\textbf{GateNet~\cite{GateNet}} & ResNet50 &  -   &  -                  & 0.889  & 0.040  & 0.885  & 0.935  & 0.034  & 0.915    \\
\textbf{$\text{F}^3$Net~\cite{F3Net}} & ResNet50 & 25.537M & 10.998G    & 0.897  & 0.035  & 0.888  & \textcolor{blue}{0.939}  & \textcolor{blue}{0.028}  & 0.917    \\
\textbf{EGNet~\cite{zhao2019EGNet}} & ResNet50 & 111.660M & 198.253G    & 0.893  & 0.039  & 0.885  & 0.938  & 0.031  & 0.918    \\
\textbf{PoolNet~\cite{Liu2019PoolSal}} & ResNet50 & 68.261M & 62.406G   & 0.894  & 0.036  & 0.886  & 0.938  & 0.030  & 0.918    \\
\textbf{MINet~\cite{MINet-CVPR2020}} & ResNet50 & 162.378M & 70.495G    & 0.888  & 0.037  & 0.884  & 0.936  & 0.029  & 0.919   \\
\textbf{CPD~\cite{Wu_2019_CVPR}} & ResNet50 & 47.850M & 11.877G         & 0.865  & 0.043  & 0.869  & 0.925  & 0.034  & 0.906   \\
\textbf{SCRN~\cite{wu2019stacked}} & ResNet50 & 25.226M & 10.086G       & 0.888  & 0.039  & 0.885  & 0.934  & 0.034  & 0.916    \\
\textbf{ITSD~\cite{zhou2020interactive}} & ResNet50 & 26.074M & 15.937G & 0.883  & 0.041  & 0.885  & 0.934  & 0.031  & 0.917   \\
\textbf{OURS} & ResNet50 & 28.838M & 8.083G                             & \textcolor{blue}{0.898}  & \textcolor{blue}{0.033}  & 0.891 & \textcolor{green}{0.940}  & \textcolor{blue}{0.028}  & 0.921    \\
\midrule

\multicolumn{10}{c}{\textbf{More light-weighted backbone}} \\
\midrule
\textbf{CSNet~\cite{hou2017deeply}} & None & \textcolor{red}{0.141M} & \textcolor{green}{1.185G} & 0.819  & 0.074  & 0.822 & 0.899  & 0.059  & 0.880    \\
\textbf{OURS} & EfficientNet-B0 & \textcolor{green}{4.606M} & \textcolor{red}{0.787G}             & 0.891  & 0.035  & 0.890 & 0.936  & 0.030  & 0.920    \\
\textbf{CSF~\cite{hou2017deeply}} & EfficientNet-B3 & 12.328M & 1.961G                           & 0.892  & \textcolor{green}{0.032}  & \textcolor{blue}{0.894}  & 0.936  & \textcolor{green}{0.027}  & 0.921  \\
\textbf{$\text{F}^3$Net~\cite{F3Net}} & EfficientNet-B3 & 12.588M & 5.701G  & \textcolor{green}{0.906}  & \textcolor{blue}{0.033}  & \textcolor{green}{0.898}  & \textcolor{red}{0.944}  & \textcolor{red}{0.025}  & \textcolor{green}{0.926}    \\
\textbf{MINet~\cite{MINet-CVPR2020}} & EfficientNet-B3 & 14.793M & 7.363G                        & 0.879  & 0.044  & 0.875 & 0.929  & 0.036  & 0.909     \\
\textbf{ITSD~\cite{zhou2020interactive}} & EfficientNet-B3 & \textcolor{blue}{11.374M} & 4.148G  & 0.894  & 0.041  & \textcolor{blue}{0.894} & \textcolor{blue}{0.939}  & 0.034  & \textcolor{blue}{0.924}    \\
\textbf{OURS} & EfficientNet-B3 & 11.522M & \textcolor{blue}{1.738G}                              & \textcolor{red}{0.907}  & \textcolor{red}{0.030}  & \textcolor{red}{0.905}    & \textcolor{red}{0.944}   & \textcolor{green}{0.027}  & \textcolor{red}{0.928}    \\
\bottomrule[1pt]
\end{tabular}%
}
\vspace{5pt}
\caption{ Quantitative comparison of our method against other SOD methods on DUST-TE and HKU-IS datasets. All these models are trained on DUTS-TR. The performances ranked first, second and third are marked by \textcolor{red}{red}, \textcolor{green}{green} and \textcolor{blue}{blue} respectively. }\label{tab:sota-comp1}
\end{table}

\begin{table}
\centering
\setlength{\tabcolsep}{0.5mm}{
\begin{tabular}{l ccc ccc ccc ccc}
\toprule[1pt]
  & \multicolumn{3}{c}{ECSSD} & \multicolumn{3}{c}{PASCAL-S} & \multicolumn{3}{c}{DUT-O} & \multicolumn{3}{c}{SOD} \\
\cmidrule(l){2-4} \cmidrule(l){5-7} \cmidrule(l){8-10} \cmidrule(l){11-13}
\textbf{Model} & \textbf{$\text{F}_{\text{max}}$$\uparrow$} & \textbf{MAE$ \downarrow $} & \textbf{S$ \uparrow $} & \textbf{$\text{F}_{\text{max}}$$\uparrow$} & \textbf{MAE$ \downarrow $} & \textbf{S$ \uparrow $} & \textbf{$\text{F}_{\text{max}}$$\uparrow$} & \textbf{MAE$ \downarrow $} & \textbf{S$ \uparrow $} & \textbf{$\text{F}_{\text{max}}$$\uparrow$} & \textbf{MAE$ \downarrow $} & \textbf{S$ \uparrow $} \\
\midrule[1pt]

\textbf{CSNet~\cite{hou2017deeply}} & 0.914  & 0.069  & 0.888  & 0.835  & 0.104  & 0.813  & 0.792  & 0.080  & 0.803 & 0.827  & 0.139  & 0.747  \\
\textbf{OURS+EffiB0}                & 0.942  & 0.038  & 0.918  & 0.872  & \textcolor{blue}{0.063}  & 0.858  & \textcolor{blue}{0.827}  & \textcolor{green}{0.052}  & 0.841  & 0.873  & 0.099  & 0.795 \\
\textbf{CSF~\cite{hou2017deeply}}  & 0.944  & \textcolor{blue}{0.034}  & 0.921  & 0.872  & \textcolor{green}{0.061}  & \textcolor{blue}{0.860}  & 0.826  & \textcolor{green}{0.052}  & \textcolor{blue}{0.844}  & 0.881  & \textcolor{green}{0.089}  & 0.808  \\
\textbf{$\text{F}^3$Net~\cite{F3Net}} & \textcolor{green}{0.947}  & \textcolor{red}{0.032}  & \textcolor{green}{0.925}  & \textcolor{red}{0.888}  & \textcolor{red}{0.058}  & \textcolor{green}{0.871}  & \textcolor{red}{0.844}  & \textcolor{blue}{0.056}  & \textcolor{blue}{0.844}  & \textcolor{green}{0.890}  & \textcolor{red}{0.083}  & \textcolor{red}{0.821}  \\
\textbf{MINet~\cite{MINet-CVPR2020}} & 0.936  & 0.043  & 0.912  & \textcolor{blue}{0.873}  & 0.070  & 0.855  & 0.813  & 0.067  & 0.821 & 0.858  & 0.101  & 0.795  \\
\textbf{ITSD~\cite{zhou2020interactive}} & \textcolor{blue}{0.945}  & 0.042  & \textcolor{blue}{0.924}  & \textcolor{green}{0.877}  & 0.065  & \textcolor{red}{0.872}  & \textcolor{green}{0.834}  & 0.058  & \textcolor{green}{0.854}  & \textcolor{blue}{0.882}  & 0.096  & \textcolor{green}{0.815}  \\
\textbf{OURS} & \textcolor{red}{0.950}  & \textcolor{green}{0.033}  & \textcolor{red}{0.927}  & \textcolor{red}{0.888}  & \textcolor{red}{0.058}  & \textcolor{red}{0.872}  & \textcolor{red}{0.844}  & \textcolor{red}{0.047}  & \textcolor{red}{0.857}  & \textcolor{red}{0.893}  & \textcolor{blue}{0.091}  & \textcolor{blue}{0.811} \\
\bottomrule[1pt]
\end{tabular}%
}
\vspace{5pt}
\caption{ Quantitative comparison of our method against other SOD methods on ECSSD, PASCAL-S, DUT-O and SOD datasets. All these models are trained on DUTS-TR. The performances ranked first, second and third are marked by \textcolor{red}{red}, \textcolor{green}{green} and \textcolor{blue}{blue} respectively. No pretrained backbone is used in CSNet. `OURS+EffiB0' indicates the variant of our method using EfficientNet-B0 as the backbone. Other methods use EfficientNet-B3 as the backbone. }\label{tab:sota-comp2}
\end{table}

\begin{figure*}[t]
\centering
\includegraphics[width=0.48\columnwidth]{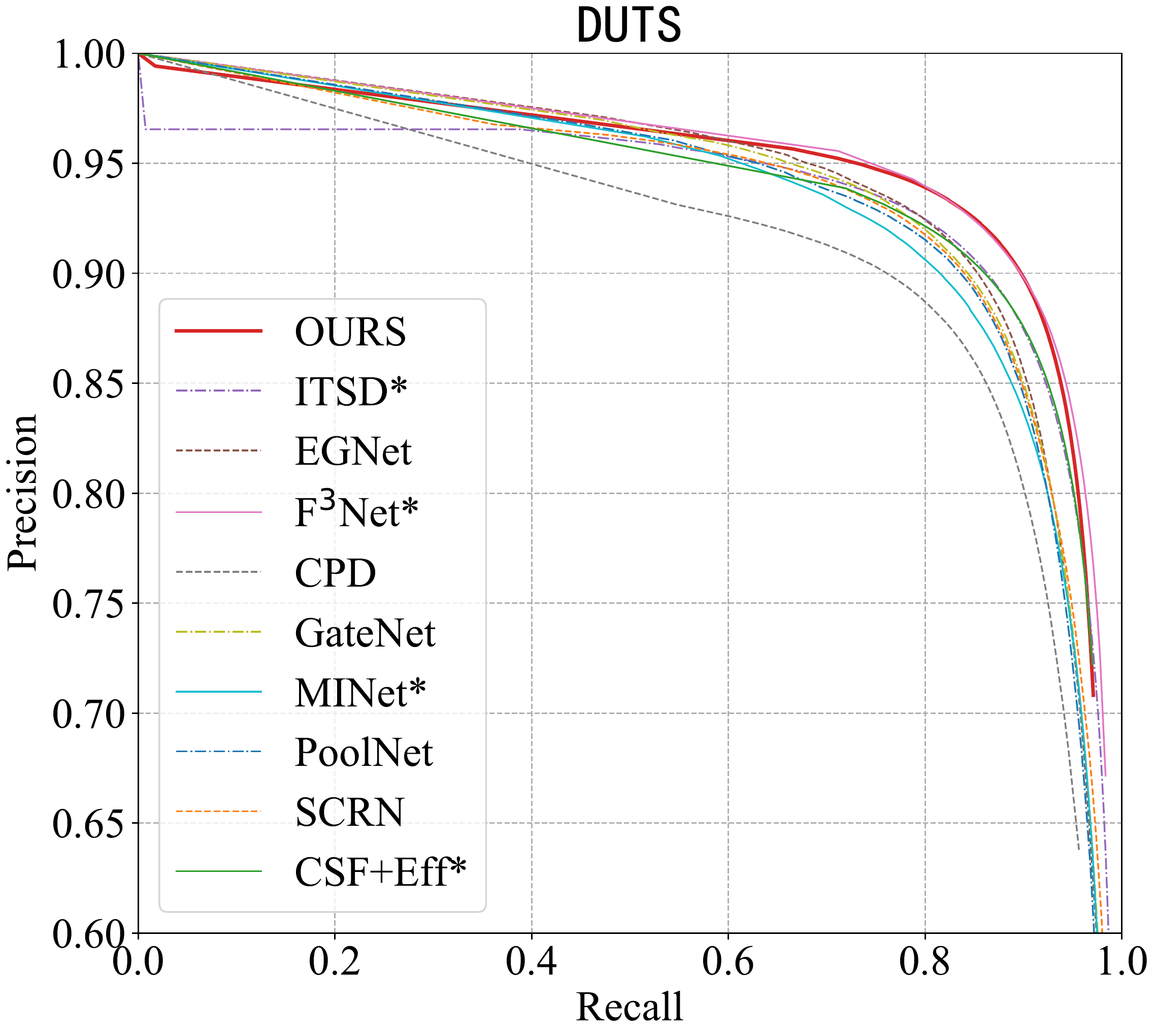}
\includegraphics[width=0.48\columnwidth]{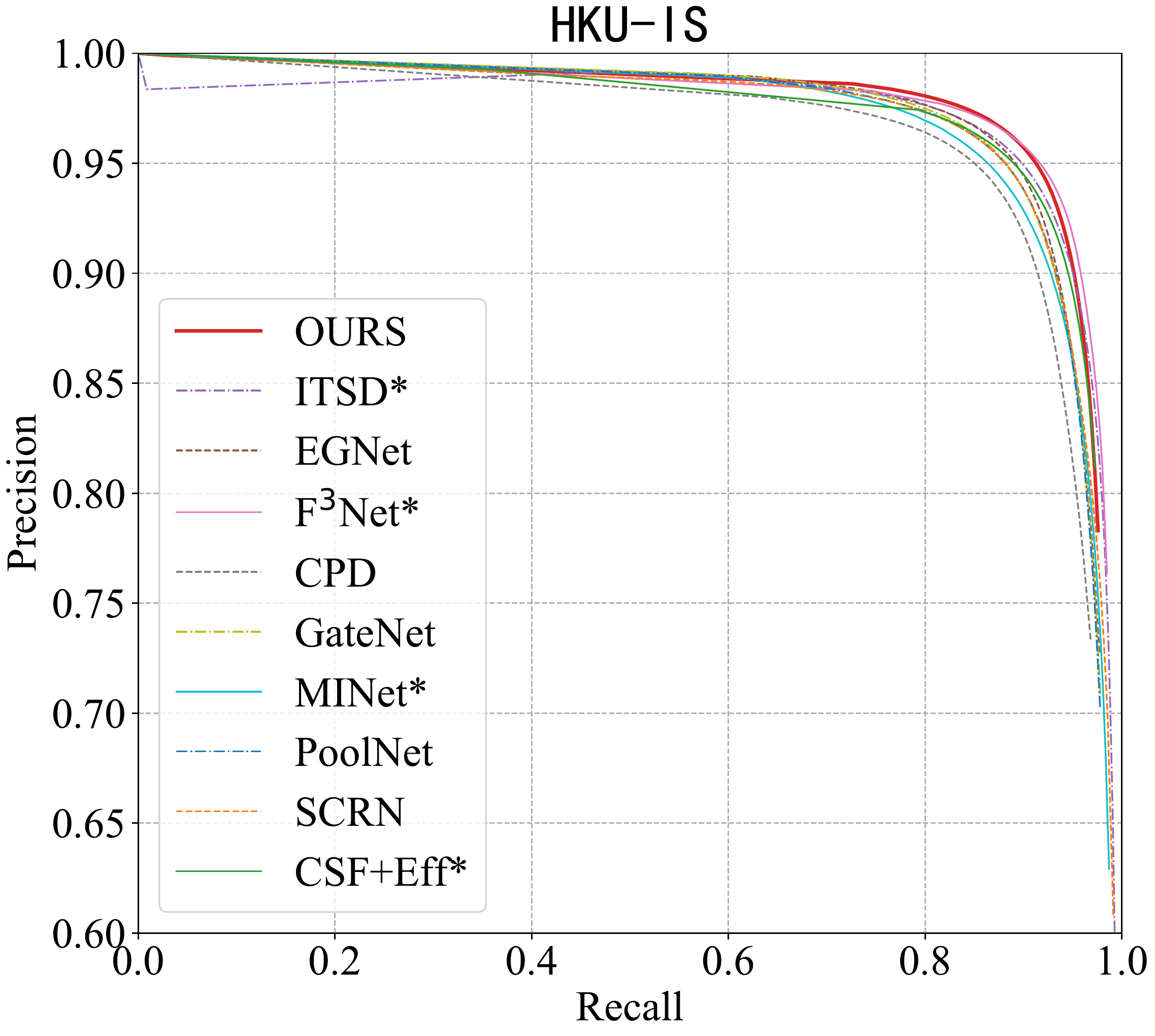}
\caption{ PR curves on the two largest salient object datasets. * denotes the models use the EfficientNet-B3 as the backbone while the rest models use the ResNet50 as the backbone. }
\label{fig:pr_curves}
\end{figure*}

\subsection{Network Training}
To make our model pay more attention to the edge of salient object, we adopt the edge weighted binary cross entropy loss~\cite{F3Net} during the training stage,
\begin{align}
\label{eq:lwbce}
L_{wbce}    &= \sum_{i=1}^{H} \sum_{j=1}^{W}  (1 + \gamma \alpha_{i,j}) \textit{BCE}\left(P_{i,j}, Y_{i,j} \right), \\
\alpha_{i,j} &=  \left\lvert \frac{ \sum_{m=-\delta}^{\delta} \sum_{n=-\delta}^{\delta} Y_{i+m,j+n} }{ (2\delta+1)^2} - Y_{ij} \right\rvert,
\end{align}
where $\textit{BCE}(\cdot,\cdot)$ is the binary cross entropy loss function, and $\gamma$ is a constant. $ P_{i,j} $ and $ Y_{i,j} $ are the value at position $(i,j)$ of $\mathbf P$ and the ground-truth saliency map $\mathbf Y$, respectively. 
$\alpha_{i,j}$ measures the weight assigned to the loss at position $(i,j)$, which receives a relatively large when $(i,j)$ locates around the boundaries of salient objects.
$\delta$ represents the radius of window size for calculating $\alpha_{i,j}$, and mirrored padding is adopted to fill positions outside the border of the image. Adam~\cite{kingma2014adam} is used to optimize network parameters.

\section{Experiments}

\begin{figure*}[t]
\begin{center}
\includegraphics[width=1\linewidth]{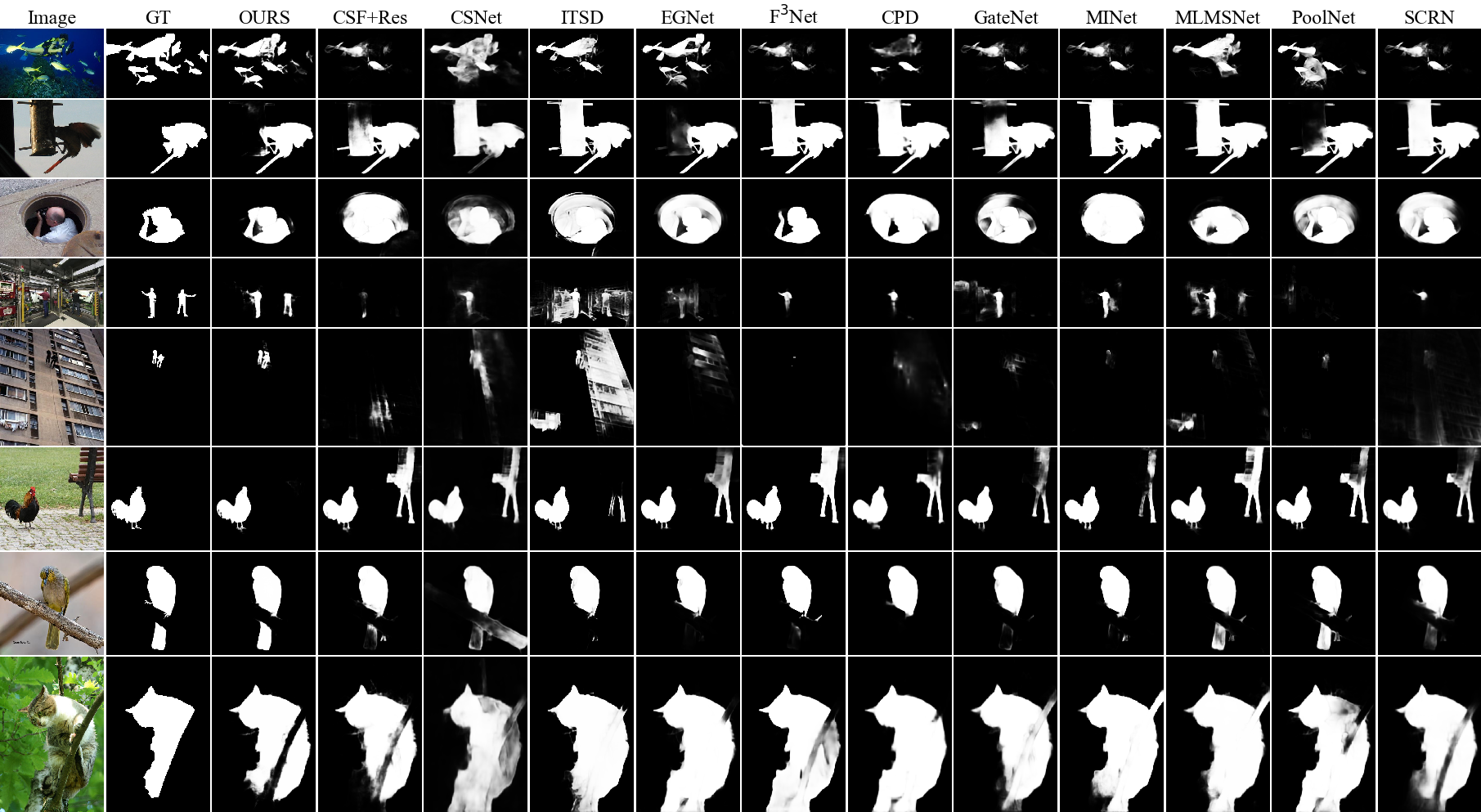}
\end{center}
   \caption{Qualitative comparison of our method against other SOD methods. }
   \label{fig:compvis}
\end{figure*}

\subsection{Datasets \& Evaluation Metrics}
The DUTS~\cite{Wang_DUTS} is the largest dataset for salient object detection, containing 10,553 training images (DUTS-TR) and 5,019 testing images (DUTS-TE).
Our proposed model is trained with images of DUTS-TR and evaluated on six commonly used salient object detection datasets, including DUTS-TE, HKU-IS~\cite{li2015visual}, ECSSD~\cite{yan2013hierarchical}, PASCAL-S~\cite{li2014secrets}, DUT-OMRON~\cite{yang2013saliency}, and SOD~\cite{movahedi2010design}.

Three metrics are adopted to evaluate the performance of SOD methods, including the maximum of F-measure ($\textrm{F}_{max} $)~\cite{achanta2009frequency}, mean absolute error ($\textrm{MAE}$), and S-measure ($\textrm{S}$)~\cite{FanStructMeasureICCV17}.

\subsection{Implementation Details}
In our experiments, our proposed top-down flow mechanism is integrated with two kinds of backbone models, including ResNet50~\cite{he2016deep} and EfficientNet~\cite{tan2019efficientnet}.
For ResNet50, we adopt the knowledge distillation strategy in~\cite{shen2020meal} to initialize network parameters. The other models, EfficientNet-B0 and EfficientNet-B3, are pretrained on Imagenet~\cite{deng2009imagenet}.
The trainable parameters of the decoder are initialized as in~\cite{he2015delving}.
Random horizontal flipping and multi-scale training strategy (0.8,0.9,1.0,1.1 and 1.2 times geometric scaling) are applied for data augmentation.
All models are trained with 210 epochs and the batch size is set as 1.
The learning rate is initially set to $1.0\times10^{-5}$ and $4.5\times10^{-4}$ respectively for ResNet50 and EfficientNet, and divided by 10 at the at the 168-th epoch. Hyper-parameters in (\ref{eq:lwbce}) are set as $\gamma=3$ and $\delta=10$. A variety of values $\{2,4,8,16,32\}$ are tested for the compression ratio $r$. Without specification, $r$ is set as 4, 2 and 2 for ResNet50, EfficientNet-B0 and EfficientNet-B3, respectively.
Our proposed model is implemented with PyTorch, and one 11GB NVIDIA GTX 1080Ti GPU is used to train all models.

\begin{figure}[t]
\begin{center}
\includegraphics[width=0.6\linewidth]{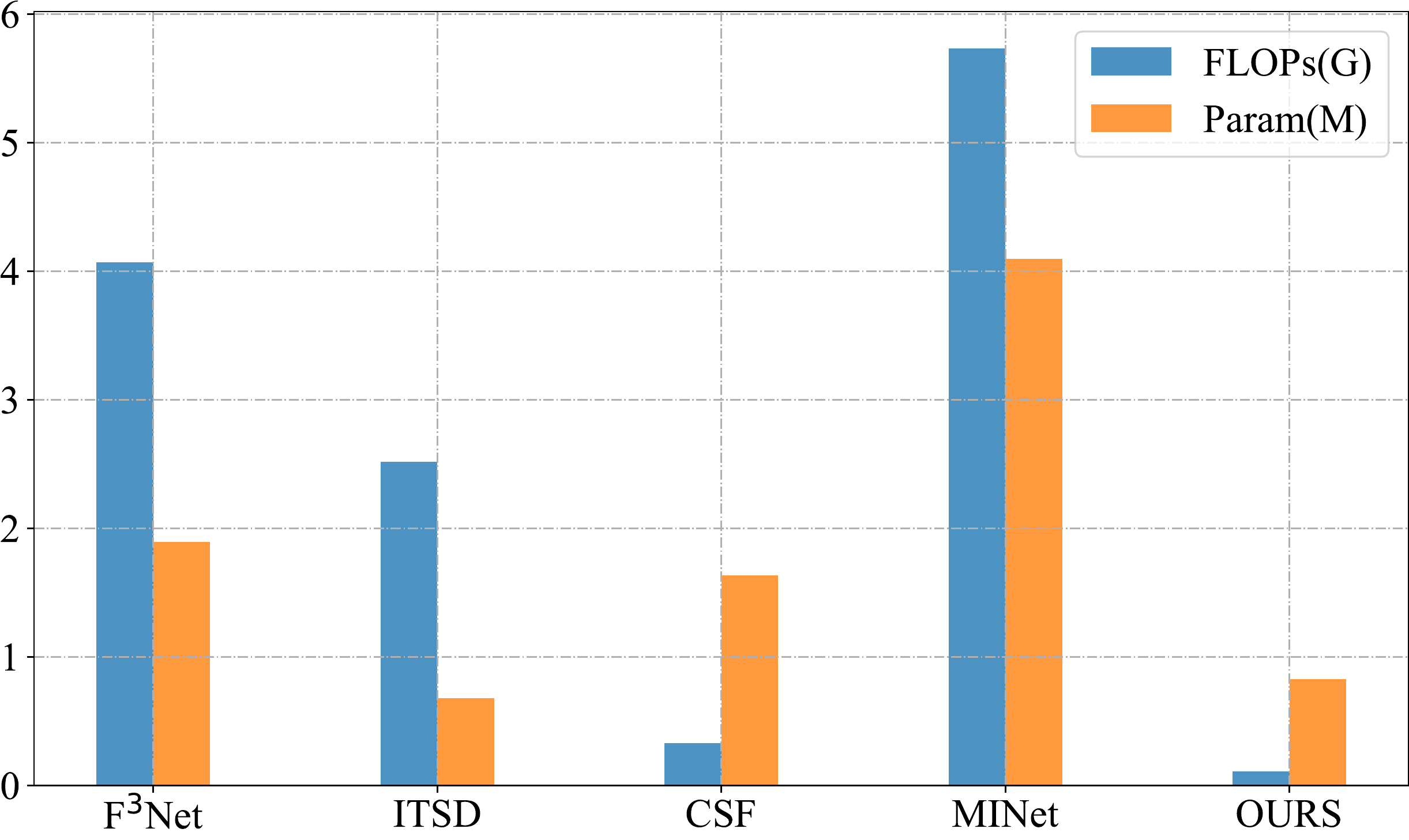}
\end{center}
   \caption{Comparisons of the parameters and FLOPs between different decoders based on EfficientNet-B3.   }
\label{fig:ResnEfficientNetet-Based}
\end{figure}
\begin{table}[t]
  	\centering
	\setlength{\tabcolsep}{5mm}{
    \begin{tabular}{cccccc}
    \toprule[1pt]
    \textbf{PCSP} & \textbf{PPM}  & \textbf{$ \text{F}_{\text{max}} $} & \textbf{MAE} & \textbf{S} \\ 
    \midrule
    \xmark & \xmark  & 0.887  & 0.039  & 0.883  \\
    \xmark & \cmark  & 0.891  & 0.037  & 0.886  \\ \hline
    1 & \cmark  & 0.891  & 0.034  & 0.890  \\
    2 & \cmark  & 0.894  & 0.034  & 0.892  \\
    3 & \cmark  & 0.896  & 0.034  & 0.891  \\ \hline
    4 & \xmark  & 0.891  & 0.037  & 0.886  \\
    4 & \cmark  & \textbf{0.898}  & \textbf{0.033}  & \textbf{0.891}  \\
    \bottomrule[1pt]
    \end{tabular}
    }
  \caption{Ablation study on DUTS-TE dataset, using backbone ResNet50.
  \cmark/\xmark indicates whether the module is used or not. The number of PCSP denotes the number of most top feature maps of the encoder which are propagated to bottom convolutional blocks of the decoder.}\label{tab:ablation}%
\end{table}%

\begin{table}[t]
  	\centering
	\setlength{\tabcolsep}{2.3mm}{
    \begin{tabular}{lccccc}
    \toprule[1pt]
   \textbf{DCB} &  \textbf{$ \text{F}_{\text{max}} $} & \textbf{MAE} & \textbf{S}  & \textbf{FLOPs} & \textbf{Param} \\
    \midrule
    3$\times$3 & 0.907  & 0.0305  & 0.905  & 0.108G & 0.825M \\

    FAM & 0.904  & 0.0312  & 0.902  & 0.123G  & 1.906M \\
    \bottomrule[1pt]
    \end{tabular}%

  }
  \caption{Inner comparisons of different variants of our method based on EfficientNet-B3.
 `DCB' indicates the convolutional block adopted in every stage of the decoder. 
  `FAM' means the module containing a single 3$\times$3 convolution operation is replaced with the FAM~\cite{Liu2019PoolSal} in every stage of the decoder.
  }\label{tab:variants}%
\end{table}%

\begin{table}[t]
  \centering
    \begin{tabular}{ccccrr}
    \toprule
    \textbf{scale} & \multicolumn{1}{c}{\textbf{$\text{F}_\text{max}$}} & \textbf{MAE} & \textbf{S} & \multicolumn{1}{c}{\textbf{Param}} & \multicolumn{1}{c}{\textbf{FLOPs}} \\
    \midrule
    32   & 0.884  & 0.036  & 0.880  & 379.50K & 63.77M \\
    16   & 0.890  & 0.035  & 0.888  & 840.92K & 154.83M \\
    8    & 0.893  & 0.034  & 0.887  & 2.01M & 420.96M \\
    4    & 0.898  & 0.033  & 0.891  & 5.33M & 1.29G \\
    2    & 0.901  & 0.031  & 0.895  & 15.90M & 4.37G \\
    \bottomrule
    \end{tabular}%

  \caption{Comparisons of performance, parameters and FLOPs which is based on ResNet50 using different compression scale. \textbf{Param} and \textbf{FLOPs} denote the parameters and FLOPs of the decoder. }\label{tab:diff_scale}%
\end{table}%
\subsection{Comparison with State-of-the-arts}

As presented in Table \ref{tab:sota-comp1} and \ref{tab:sota-comp2}, we compare our method against various existing SOD methods. For a fair comparison, we reimplemented very recently proposed SOD algorithms, including CSF~\cite{gao2020highly}, F$^3$Net~\cite{F3Net}, MINet~\cite{MINet-CVPR2020} and ITSD~\cite{zhou2020interactive}, via replacing their original backbones with EfficientNet-B3.
FLOPs are calculated with a $ 288 \times 288 $ input image.
Table \ref{tab:sota-comp1} presents experimental results of various SOD methods which use VGG~\cite{russakovsky2015imagenet}, ResNet, and EfficientNet as backbones, on the two largest datasets, DUTS and HKU-IS.
Our method surpasses state-of-the-art methods while consuming much fewer FLOPs. 
When using ResNet50 as backbone, our method achieves marginally better performance than the second best method F$^3$Net~\cite{F3Net}, and the FLOPs consumed by our method are 2.915G fewer.
When integrated with EfficientNet-B3, on the DUTS-TE dataset, the S-measure produced by our method is 0.007 higher than that produced by F$^3$Net, while the FLOPs consumed by our method are 30.48\% of those consumed by F$^3$Net.
Table \ref{tab:sota-comp2} showcases SOD performance on the other 4 datasets, including ECSSD, PASCAL-S, DUT-O, and SOD.
On the DUT-O dataset, our method gives rise to results having 9.62\% lower MAE, compared to the results of CSF.
Without using a pre-trained backbone, the performance of CSNet is inferior though it costs a small number of parameters and FLOPs.
On the basis of EfficientNet-B0, our proposed method contributes to a very efficient SOD model which has less FLOPs than CSNet   while maintaining appealing SOD performance. Overall, our method achieves the best performance across backbone models.

In addition, We follow \cite{zhao2019EGNet} to compare the precision-recall curves of our approach with the state-of-the-art methods on the DUTS and HKU-IS datasets. 
A gallery of SOD examples is also visualized in Figure \ref{fig:compvis} for qualitative comparisons. Our method performs clearly better than other
methods, across small and large salient objects.

\subsection{Ablation Study}\label{sec:inner-comp}
\noindent \textbf{Efficacy of Main Components} In this experiment, we first verify the efficacy of main components in our proposed model, including the progressive compression shortcut path (PCSP) and the PPM for global feature extraction.
ResNet50 is used to construct the backbone of our proposed model and SOD performance is evaluated on the DUTS-TE dataset.
The experimental results are presented in Table \ref{tab:ablation}.
As more top-down feature maps are used to complement high-level semantic information in bottom convolution layers via multiple PCSPs, the performance of our method increases consistently. The adoption of 4 PCSPs, induces to performance gains of 0.005 (when PPM is not used) and 0.011 (when PPM is used) on the $\text{F}_{\text{max}}$ metric.
Besides, we can observe that the global information provided by the PPM and high-level semantic information provided by the PCSP can complement each other. Without using any of the two modules, performance degradation is caused.

\noindent \textbf{Different Variants of Our Method} We further provide inner comparisons between variants of our proposed model, in Table \ref{tab:variants}.
To validate whether more complicated convolution blocks is effective in the decoder of our method, we replace the 3$\times$3 convolution layer with the FAM block used in~\cite{Liu2019PoolSal} to build the fusion module of each decoding stage. However, no obvious performance gain is obtained.

\subsection{Efficiency Discussions}
\noindent \textbf{Analysis of Compression Ratio} The influence of using different ratios to compress the features of the encoder as in (\ref{eq:compress})  is illustrated in Table \ref{tab:diff_scale}.
As discussed in Section \ref{sec:light}, there are large amounts of redundant information in the features extracted by the backbone since it is pre-trained for image recognition. Hence, using a moderately large ratio (up to 16) to compress features of the network backbone has no significant effect on the SOD performance, according to the results reported in Table \ref{tab:diff_scale}. The benefit of using a large compression ratio is achieving the goal of light-weighted network design in our method while not causing unbearable performance decrease.

\noindent \textbf{Complexity of Decoder} As shown in Figure \ref{fig:ResnEfficientNetet-Based}, the decoder of our proposed model costs significantly fewer FLOPs than the decoders of recent SOD models, including F$^3$Net, ITSD, CSF and MINet. The parameters and FLOPs are counted using the backbone of Efficient-B3. As can be observed from Table \ref{tab:sota-comp1} and \ref{tab:sota-comp2}, our method outperforms these methods on most datasets and metrics. This indicates that our method achieves better performance even if less memory is consumed.

\section{Conclusion}
In this paper, we first revisit existing CNN-based top-down flow architectures. Then, to make full usage of multi-scale high-level feature maps, progressive compression shortcut paths are devised to enhance the propagation semantic information residing in higher-level features of the encoder to bottom convolutional blocks of the decoder, which form the novel densely nested top-down flows. Extensive experiments on six widely-used benchmark datasets indicate that the proposed SOD model can achieve state-of-the-art performance. Notably, the computational complexity and model size of the proposed framework are also very light-weighted.







\end{document}